\documentclass[mathpm,proof]{bioinformatics-altered}

\usepackage{amssymb,amsfonts,url,times,epsfig}
\def\email#1{#1}

\begin{document}


\verso{A.S.Yeh \etal}
\recto{Evaluation of text-mining to aid database curation}

\title{Evaluation of text data mining for database curation:
lessons learned from the KDD Challenge Cup} 

\author{Alexander S.\ Yeh}
\author{Lynette Hirschman}
\author{Alexander A.\ Morgan}


\address{The MITRE Corporation,
202 Burlington Road,
Bedford, MA 01730,
USA}

\maketitle

\renewcommand{\thefootnote}{\arabic{footnote}}

\begin{abstract}
\begin{subabstract}[Motivation]
{\begin{picture}(0,0)
\put(-52,215){In {\it Bioinformatics} Vol. 19 Suppl. 1 2003, pages i331-i339. http://bioinformatics.oupjournals.org/cgi/reprint/19/suppl\_1/i331}
\put(-52,203){This is close to how it appears on the publisher's website. The article wording is the same.}
\put(-52,-526){cs.CL/0308032}
\end{picture}}
The biological literature is a major repository of
knowledge. Many biological databases draw much of their content from a
careful curation of this literature. However, as the volume of
literature increases, the burden of curation increases. Text mining
may provide useful tools to assist in the curation process. To date,
the lack of standards has made it impossible to determine whether text
mining techniques are sufficiently mature to be useful.
\end{subabstract}
\begin{subabstract}[Results]
We report on a Challenge Evaluation task that we
created for the Knowledge Discovery and Data Mining (KDD) Challenge
Cup. We provided a training corpus of 862 articles consisting of
journal articles curated in FlyBase, along with the associated lists of
genes and gene products, as well as the relevant data fields from
FlyBase. For the test, we provided a corpus of 213 new
(`blind') articles; the 18 participating groups provided systems 
that flagged articles for curation, based on whether the article contained 
experimental evidence for gene expression products.  We report on the 
the evaluation results and
describe the techniques used by the top performing groups.
\end{subabstract}
\begin{subabstract}[Contact]
\email{asy@mitre.org}
\end{subabstract}
\begin{subabstract}[Keywords]
text mining, evaluation, curation, genomics, data management
\end{subabstract}
\end{abstract}
\baselineskip12pt plus .2pt minus .05pt

\section{Introduction}
The research literature is a major repository of biological
knowledge. To make this knowledge accessible, it is
translated by expert curators into entries in biological databases.
This serves several purposes: experts consolidate
data about a single organism or a single
class of entity (e.g., proteins) in one place, often in
conjunction with sequence information. Second, this process makes the
information searchable through a variety of automated techniques,
given that the curators use standardized terminologies or ontologies.
However, it is becoming more and more difficult for curators to keep up
with the increasing volume of literature, creating a demand for automated 
curation aids.

There has been a growing volume of work in text mining for
biological literature, but until now, there has been no way to compare
results of the different systems \citep{Hirschman02}.  Several related
fields have addressed this problem by organizing open `challenge'
evaluations, e.g., for protein structure
prediction, there have been the successful CASP
evaluations (Critical Assessment of Techniques for Protein Structure
Prediction, http://predictioncenter.llnl.gov/). For natural language processing, 
there was the series of Message
Understanding Conferences (MUCs) for information
extraction on newswire text \citep{Hirschman98}. The Text REtrieval 
Conferences (TREC, http://trec.nist.gov/; \citep{Voorhees02}) 
for information retrieval are ongoing.

The idea behind these series of open evaluations has been to attract
teams to work on a problem by providing them with real (or realistic)
training and test data, as well as objective evaluation metrics. These
data sets are often hard to obtain, and the open evaluation makes it
much easier for groups to build systems and compare performance on a
common problem.  If many teams are involved, the results are a measure
of the state-of-the-art for that task. In addition, when the
teams share information about their approaches and the evaluations are
repeated over time, then the research community can demonstrate
measurable forward progress in a field.

For these reasons, we decided that it would be valuable to test
whether text mining techniques could help the curation process. To do
this, we created and ran a common challenge evaluation (a contest)
using data from a biological database and a task performed by curators
of biological databases.
The contest that we created and ran was Task 1 (of 2 tasks) of the KDD
Challenge Cup 2002, a competition held in conjunction with the ACM
SIGKDD International Conference on Knowledge Discovery and Data Mining
(KDD), July 23-26, 2002.\footnote{See 
http://www.biostat.wisc.edu/$\sim$craven/kddcup/ for a description of the 
task.} This contest focused on text mining to
provide semi-automated aids for biological database curation \citep{Yeh03}. FlyBase,
a publicly available database on Drosophila genetics and molecular
biology \citep{FlyBase02}, supplied the data and biological expertise.
This paper describes the results and lessons that we learned from
setting up and running this contest.

\section{Methods: Contest Set-Up}
For this contest, we drew on the work performed by Prof.  William
Gelbart and colleagues at Harvard in connection with FlyBase Harvard.
We discussed how to provide automated aids for curating biomedical databases 
with the
FlyBase curators and settled on a fundamental task at the beginning of
the FlyBase Harvard curation pipeline, that of identifying the
papers to be curated for Drosophila gene expression information.

FlyBase Harvard curates journal articles containing experimental gene expression
evidence, specifically, experimental
evidence about the products -- mRNA transcripts (TR) and
proteins/polypeptides (PP) --  associated with a given gene.

We defined the following task for the contest, based on materials
obtained from FlyBase:
\begin{itemize}
 \item Given a set of papers (full text) on genetics or molecular
biology and, for each paper, a list of the genes mentioned in that
paper:

 \item Determine whether the paper meets the FlyBase gene-expression
curation criteria, and for each gene, indicate whether the full paper
has experimental evidence for gene products (mRNA and/or protein).
\end{itemize}
For each paper, a system needed to return three things:
\begin{enumerate}
 \item A ranked list of papers in order of probability of the
need for curation, where papers containing experimental evidence of
interest should rank higher than papers that did not contain such evidence;

 \item A yes/no decision on whether to curate each paper;

 \item For each gene in each paper, a yes/no decision about whether
the paper contained experimental evidence for the gene's products
(RNA and protein/polypeptide).
\end{enumerate}

The KDD Challenge Cup schedule included a 6 week period when the
training data was made available for the contestants to build
and train a system, followed by a two week period to
complete the running of the test material. The results were then submitted to 
MITRE for final scoring.

\subsection{The training and test data}\label{ss:train-and-test-data}
The training data set consisted of 862 `cleaned' full text papers,
of which 283 had been judged to need curation. Each paper came to the
Harvard curators with a list of the genes (in a standardized
nomenclature) mentioned in the paper. Along with its standardized
nomenclature, the FlyBase database provides synonym lists for each
gene. These resources, along with the set of
relevant FlyBase database entries for each paper, were provided 
to KDD participants as
part of the training data. 

The test data set came from papers that had already been curated for
genes (so the gene list was available to both the FlyBase gene product
curators and the general public), but for which the gene product
curation was not public yet. In the end, the test set
consisted of 213 papers, together with the genes mentioned in each
paper.

For each paper, we `cleaned' the text by converting non-plain text
(superscripts, subscripts, italics, Greek letters) into plain text;
this was critical to preserve distinctions in meaning for further
processing.  For example, in `{\it Appl\/}$^{\rm d}$', the {\it
Appl\/} in italics indicates that the Appl gene is being mentioned
(and not the protein) and the superscript {\it d} indicates that the
Appl gene's {\it d} allele is being mentioned.  The resulting
conversion produced the form `@Appl@[d]', using the conventions
developed by FlyBase for their gene name lists.

The list of genes for a paper was given in the form of a template in
XML that also indicated the yes/no decisions to be made. For the
training papers, a filled-out version of this template was also
provided. For example, the following template indicated the mention of
the \verb+sws+ and \verb+Ecol\lacZ+ genes in the associated paper:

\begin{verbatim}
<curate>?</curate>
<gene symbol="sws">
         <tr>?</tr><pp>?</pp></gene>
<gene symbol="Ecol\lacZ">
         <tr>X</tr><pp>X</pp></gene>
\end{verbatim}

Systems gave their yes/no (\verb+Y+/\verb+N+) answers by
returning these templates with the \verb+?+'s replaced by \verb+Y+ or
\verb+N+. For each gene, returning \verb+<pp>Y</pp>+ meant that a
system found experimental data of interest in the paper for some
{\em protein} of that gene. Returning \verb+<pp>N</pp>+ meant that a system
did not; \verb+<tr>Y</tr>+ and \verb+<tr>N</tr>+ indicated analogous
findings for that gene's {\em transcripts}.

Lethal (e.g., \verb+l(2)52A+), foreign (e.g., \verb+Ecol\lacZ+) and
anonymous (e.g., \verb+anon-56Cb+) genes were especially hard to
handle and were deemed `special'. Systems did not have to answer
\verb+Y+/\verb+N+ for those genes' products. We indicated this by
replacing the appropriate \verb+?+'s with \verb+X+'s.

The overall decision on whether a paper had experimental evidence for
a product of any gene ({\em including} `special' genes) was
indicated by changing the \verb+?+ in \verb+<curate>?</curate>+ into a
\verb+Y+ for yes and \verb+N+ for no.

For the training papers, we also provided the experimental data that
FlyBase extracted from that paper. For example, below is relevant evidence
from \citet{Kolhekar97}:
\begin{verbatim}
(gene="Phm" product="Phm-P1" 
 ptype="pp" evtype="asm"): 
                  immunolocalization
\end{verbatim}
This indicated that the assay mode ({\em asm}) was 
{\em immunolocalization},
used on the {\em Phm-P1} protein product ({\em pp}) of the {\em Phm} gene.

The data sets presented a number of complications.  First, the list of synonyms 
for the genes provided to the contestants was
not complete because of the many typographical variants of names. For example,
Flybase listed the following 7 synonyms for the {\it Appl} gene:
\begin{center}
\begin{tabular}{lllll}
{\tt APPL}, & {\tt appl}, & {\tt EG:65F1.5}, & {\tt CG7727}, & {\tt 
BcDNA:GH04413},\\
\multicolumn{5}{l}{{\tt \&bgr; amyloid protein precursor-like},} \\
\multicolumn{5}{l}{{\tt \&bgr;-amyloid-protein-precursor-like}.}
\end{tabular}
\end{center}
But this list did not include 
the synonym {\it APP-like} as a gene name, which appears in \citet{Rosen89}.
In addition, some names are not unique to a particular gene. For
example, {\it Clk} is both a symbol for the {\it Clock} gene and
a synonym for the {\it period} gene.
This meant that it was not trivial to map between the genes listed for an 
article and their mentions in the text.

The training set came from papers that were already curated and
publicly available from FlyBase. One small source of noise in the
training data was due to the fact that, not surprisingly, curation
standards change over time and differ between individuals. For
example, FlyBase is only interested
in gene expression results that are applicable to `regular' flies
found in the wild (wild-type), and not in expression results that 
apply just to laboratory induced mutations. In addition, FlyBase is
normally interested in wild-type experimental gene expression results
that are repeats of results found in other, earlier papers. However,
if the focus of a paper is not on the gene products in wild-type
flies, but the paper does have a few experimental results on wild-type
flies (usually to serve as controls in an experiment) that have
already been seen elsewhere, then FlyBase is {\em not} interested in
that particular paper's gene expression results. The border between a
`few' results (not of interest) and enough results to be of interest
is a bit fuzzy. Such borderline papers were removed
from the test data, but were left in the training data.

Also, it took significant reverse engineering to determine how the
experimental evidence was encoded in the database, and exactly what
kinds of information constituted experimental evidence. This reverse
engineering was not perfect, and the imperfections form another source
of residual noise. Many FlyBase transcript and protein data fields
contain experimental data, such as {\em transcript length} and {\em
assay mode}, but many others do not, including the fields {\em public protein
symbol} and {\em synonyms for transcript symbol}. There are also
fields that contain data that used to be of interest to the gene
product curators, but are no longer, because another group is now
curating these fields. Examples are {\em protein domains} and {\em
protein characteristics}. The {\em comment} field is a special case by
itself. It usually contains experimental results of interest to the
gene product curators, so we included its existence as an indicator of
a training paper having results of interest for an associated gene's
transcript or protein. But the field is used for any
information that will not fit anywhere else in FlyBase, and so the
field sometimes contains material that is either not of interest
or of borderline interest.

We originally wanted a contestant's system to provide evidence for 
its response, by indicating a passage in
the text describing relevant experimental results.  
When using a system to aid in curation, providing
such a passage would give a person checking the system a basis on
which to accept or reject that finding. But while FlyBase stores the
results of interest found in a paper, it does not indicate which
passage(s) in that paper support or describe those results.   Furthermore,
the entry
in FlyBase often uses wording that is very different from what is
explicitly stated in the passage(s).  For example, FlyBase's assay
field for the PHM protein in the paper \citep{Kolhekar97} uses the
controlled vocabulary term {\em immunolocalization}. In that paper,
there is {\em no} mention of the term `immunolocalization' (or any
similar term) in the text. Instead, the supporting text describes the
various steps taken to perform an {\em immunolocalization} assay (use
an anti-body to stain some tissue and then look at it), as illustrated
in this figure caption excerpt from \citet{Kolhekar97}:
\begin{quotation}
\noindent
{\it Figure 12.} {\bf Top.} Whole-mount tissue staining using an
affinity-purified anti-PHM antibody in the CNS and in non-neural
tissues. {\it A,} The third instar larval CNS exhibits distributed
cell body and neuropilar staining. This view displays only a portion
of the CNS; ...
\end{quotation}
Another example is that for the paper \citet{Tingvall01}, FlyBase
records that mRNA transcripts of certain reporter constructs (a
construct is a combination of a reporter gene and a gene of interest)
appear in certain parts of the body. The paper itself never explicitly
mentions any transcript. Instead, the supporting text mentions where
the associated reporter protein is detected. The FlyBase curators
infer the transcripts' locations from the places where the protein is detected.
Manually finding such `evidence' passages for use in training a
system would have been both time consuming and
difficult\footnote{Especially since it can require a lot of biology
knowledge and our contestants had more of a data-mining background
than a biology background.}.  So we dropped the passage finding
requirement.

We also originally wanted the participating systems to generate the names of the 
gene product(s) that had experimental results in a paper.  However, different 
proteins in the
FlyBase database are named using different conventions (and likewise
for transcripts). For example, FlyBase lists 5 different forms of
proteins for the {\bf Doa} gene, which are named using 4 different
conventions:
 \mbox{\bf Doa$^{+}$P105kD} is named after the form's size (105
 kilodaltons).
 \mbox{\bf Doa$^{+}$P517} is named after the form's length (517
 amino acids).
\mbox{\bf Doa-P1} and \mbox{\bf Doa-P2} are named using more recent naming conventions 
for distinct forms of {\bf Doa} protein found in the literature.
\mbox{\bf Doa$^{+}$P} is a name used for results that apply to
 one
 or more forms of {\bf Doa} protein, but the curator cannot tell from
 the
 paper which specific form(s).

Furthermore, the product names used in the papers do not always
match the corresponding FlyBase names. Determining the correspondences
may not be so difficult with FlyBase names that contain some product
property like size or length, for example `105-kD
protein' or `105-kD DOA isoform' in \citep{Yun00}, which
are mentions of \mbox{\bf Doa$^{+}$P105kD}. However, determining the
correspondences to other FlyBase product names is difficult. For
example, in the same paper, phrases like `55-kD DOA protein' and `55-kD
isoform' are recorded in FlyBase as \mbox{\bf Doa-P2}. Also in that paper,
phrases like `protein kinase', `DOA kinase', `DOA protein', and
`DOA' can either refer to all forms of DOA protein, the two forms
studied in detail in that paper (\mbox{\bf Doa$^{+}$P105kD} and
\mbox{\bf Doa-P2}) or to one or more forms, but the paper is unclear
(to a biologically-trained curator) as to which, leading the
curator to use \mbox{\bf Doa$^{+}$P}.

In addition, difficulties in determining the correspondences can lead
to difficulties in determining when a transcript or protein
described in a paper is actually new to FlyBase, and has yet to be
listed in the database. For these reasons, we avoided the issue of
naming gene expression products; we simply required the systems to provide
a `yes/no' answer for whether a paper had experimental results for that
gene's transcripts and proteins.\footnote{
Even with this simplification, products of `special' (lethal,
foreign and anonymous) genes can be hard to handle, so we added the further 
simplification that contestants did not
need to make `yes/no' decisions about these products, as mentioned earlier.}

\subsection{Scoring measures}
The contest task was divided into 3 sub-tasks.  The ranked-list and
`yes/no curate paper' sub-tasks are two possible ways to help a
curator with filtering out the papers that have no information of
interest. The ranked-list can help by providing an ordering on the
relative likelihood of a paper being of interest. If accurate,
the `yes/no curate paper' decisions are direct indicators of what
papers to concentrate on. The third sub-task (`yes/no' for products
of each gene) is a way to tell a curator what gene(s) to concentrate
on in a paper.

After defining the task and preparing the training and test data, we
developed a simple scoring method for each of the three sub-tasks. 

For the ranked-list sub-task, we used as a metric the area under the
receiver operating characteristic curve (AROC); the ROC curve
\cite[Sec.~2.8.3]{Duda01} measures the trade-off between sensitivity
(recall) and the probability of a false alarm. As the area under the
curve increases, a system will on average be more sensitive for
the same false alarm rate.

For the yes/no curation decisions for the set of papers, we used the
standard balanced F measure, which is a combination (the harmonic mean)
of {\em recall} and {\em precision}.\footnote{The balanced F measure is
\mbox{(2*{\em precision}*{\em recall})/({\em precision} + {\em
recall})}.}  {\em Recall\/} is the percentage of the correct `yes'
decisions that are actually returned by the system. This measures how
sensitive a system is in finding what it should find.  {\em Precision}
is the percentage of the `yes' decisions returned by the system that
are actually correct. This measures how specific a system is in just
finding what it should find.  

We also used the balanced F measure for the yes/no decisions on
experimental evidence for products of the genes mentioned in the
papers. The sum of these three scores (equally weighted) was used to
provide an overall system score.

\section{Results}
Overall, 18 teams returned 32 separate submissions for evaluation (up
to~3 per team). 
There were eight countries represented, including Japan,
Taiwan, Singapore, India, Israel, UK, Portugal and USA. There were
groups from industry, academia and government laboratories, often
teamed.  The top performing team, ClearForest and Celera, obtained
both the highest overall score and the highest score on the each
sub-task. 
The results of the 32 submissions for the three
metrics and the overall score (normalized to 100\%) are given 
in Table~\ref{t:results}.
The top 5 teams for the ranked-list sub-task all had close scores for
this sub-task (81\%-84\%).  
\begin{table}[t]
\begin{center}
\begin{tabular}{|l|r|r|r|r|} \hline
Sub-Task       & Best & 1st Quartile & Median & Low \\ \hline
Ranked-list:       &    84\% & 81\% & 69\% & 35\%\\
Y/N paper:  & 78\% & 61\% & 58\% & 32\% \\
Y/N products: & 67\% & 47\% & 35\% & 8\%\\ \hline
Overall: & 76\% & 61\% & 55\% & 32\% \\ \hline
\end{tabular}
\caption{Results of the 32 submissions}\label{t:results}
\end{center}
\end{table}

\subsection{High-performing teams}
We declared a winning team and three honorable mention teams. The teams
used a variety of approaches.
The winning team \citep{Regev03} used an information extraction
approach with manually constructed rules that matched against patterns
deemed of interest. A focus was finding patterns in figure captions.
These often involved linguistic constructs, such as 
noun phrases (e.g., `the developing midgut') and verb phrases (e.g., `does
not antagonize'). The output of the rules 
was combined to produce scores at both the document and gene level.

One honorable mention team was from three Singapore-based organizations
\citep{Shi03}. Their system looked for the presence of certain manually
chosen `keywords'.\footnote{A `keyword' could actually be more a
single word, e.g., {\it northern blot}.} Within each paragraph of a
paper, it computed
the distance (measured as the number of sentence boundaries
crossed) between each keyword mention and each mention of a gene name
or synonym. For each gene and keyword pair, the minimum
distance was noted, as was the number of occurrences with that minimum
distance.  The effects of different keywords on decisions about a
gene's products were combined using Na\"{\i}ve Bayes
\cite[Sec.~2.11]{Duda01}.

The honorable mention team from Imperial College and Inforsense
\citep{Ghanem03} had a system that used regular expressions\footnote{Many text pattern matching
systems use regular expressions to define the patterns, including the
{\it Perl} programming language and the {\it Unix grep} utility.} to find
particular patterns of words.  It automatically extracted these patterns 
from sentences in the training corpus.
The patterns were restricted to be
within a sentence or neighboring sentences, and to
contain gene name(s) or keyword(s) that appeared in
the experimental database fields from FlyBase associated with the training papers.
When searching for the products of a particular
gene, only sentences related to that gene were examined. The patterns
served as features to be combined by a support vector machine (SVM)
classifier \cite[Sec.~5.11]{Duda01} (http://svmlight.joachims.org),
which made the final decisions.

The honorable mention team from Verity and Exelixis (B.Chen, personal
communication) also had a system that used regular expressions and
SVMs (two types: transductive and inductive). The system ignored
certain sections of papers.

One thing these highly-ranked teams had in common is that they all
moved away from the `bag of words' approach common in text
classification and information retrieval. This approach represents a
document as an unordered  bag of words, thus losing any
grammatical relations among words.  The words are then weighted by
frequency to create a vector for each document.
These vectors are then compared to find similar documents
or passages.\footnote{The SMART information retrieval system
uses this `bag of words' approach \cite[Ch.~4]{Salton83}. Often in this approach, words are stemmed (e.g., removal of plural {\it s}) 
and stop words are removed, e.g., {\it a, of, the, on, in}. Then each document 
or passage is represented as a vector of words, generally using a variant of the 
`tf-idf' scheme, which weights words (terms) by their frequency within a document and by the inverse of 
the number of documents containing that word (inverse document
frequency). Two documents are often compared by taking an inner (dot)
product of their vectors, also known as a cosine measure.}
One group
\citep{Ghanem03} in fact tried this approach at first, but found that
the resulting system did not perform well.
In general, use of pattern matching and local context seemed to work better,
probably because it was important to associate experimental results with 
specific relevant genes; document level association may simply be too weak
for this set of tasks.

Many of the submissions came from teams, and these teams often
included biologists in the role of `domain expert'.
The domain experts seemed to be most useful for these teams
near the start of the contest. This was the indication that we got in
talking to a member of the winning team. The two honorable mention
teams who wrote descriptions of their work, \citep{Shi03} and
\citep{Ghanem03}, both mention using domain experts to produce some of
the feature lists that they used in their experiments. However, one thing
to keep in mind is that as
mentioned in {\bf The training and test data} Section, we made
several simplifications to this competition to make it less dependent
on domain knowledge.

\subsection{Test-set paper analysis}
In our post-competition analysis, we looked at several factors that
might have contributed to overall task difficulty. The first factor
was how well the training data and test data sets were matched. The
training data had 33\% of articles that were judged to contain
curatable experimental evidence for gene products. By contrast, the
test set had a statistically significantly higher 
percentage:\footnote{Significant at
the 0.005 level using a single-sided equal-variance $t$ test.} 91 papers (43\%) of
the 213 test papers had results of interest.

This led us to look at whether systems had been overly conservative in
marking a paper as containing evidence for curation; we concluded that
they had been.  Overall, 26 (81\%) of the 32 submissions marked less
than 91 test papers with `yes, curate'.\footnote{This is
statistically significantly higher (at the 0.015 level) than 50\%, the
highest expected figure if overall, the submissions were not
conservative. A 1-sided test with a Normal approximation of a binomial
distribution plus the Yates correction was used. This statistical
significance holds even if one assumes only 18 of the submissions are
independent (1~independent submission per team).}

We also tried to characterize what made a curation decision harder for an
individual paper. To do this, we counted how many of the 32
submissions made the correct `Y/N curate' decision for a given
paper; we call this number the $r'$ value for the paper.

Given the conservativeness observed above, it is not surprising that
papers which had no results of interest (correct answer marked `no')
tended to be easier than papers with results (correct answer marked
`yes'). The `no' papers had a higher average $r'$
($\overline{r'}$) than the `yes' papers.\footnote{\hspace{1ex}The
$r'$ standard deviations ($sd(r')$) are 14\% and
26\%, respectively. The difference in the averages is statistically
significant (at the 0.0005 level) using a single-sided equal-variance
$t$ test.} Another way to view this is that a larger fraction of
the `no' papers were correctly marked by over half the submissions
(had \mbox{$r' > 50\%$}) than the `yes' papers. See Table~\ref{t:yes-vs-no}.
\begin{table}[t]
\begin{center}
\begin{tabular}{|l|l|l|} \hline
Paper Type & average $r'$ & Fraction with $r' > 50\%$ \\ \hline
`no'     & 24.3 (76\%)  & 93\% (114 of 122 papers) \\
`yes'    & 17.6 (55\%)  & 54\% (49 of 91 papers)\\ \hline
\end{tabular}
\caption{`no' versus `yes' papers}\label{t:yes-vs-no}
\end{center}
\end{table}

We did a further analysis to see if we could determine what made the
`yes-curate' papers hard. We noted that all but one of the `yes'
papers (90/91) had results of interest for at least one
`regular'\footnote{\hspace{1ex}A `regular' gene is one that is not `special'
(anonymous, lethal or foreign) as mentioned in {\bf The training and test data}
Section.} gene product.  These 90 papers
could be divided into two groups.\footnote{\hspace{1ex}Products of `special'
genes were ignored in the determination of the groups' members.}
Papers in the first group had results of interest on {\em both}
transcripts and proteins. All test set papers of this type also had at
least one `regular' gene for which both transcript and protein
results were present in the paper.  Papers of the second type had results of
interest on {\em either} transcripts or proteins, but {\bf not} both. 
The {\em both} papers were easier to identify than the {\em either}
papers, with the former having a higher $\overline{r'}$ than the
latter,\footnote{\hspace{1ex}$sd(r')$ is 20\% and 16\% respectively. The difference in the averages
is statistically significant (at the 0.0005 level) using a
single-sided equal-variance $t$ test.} as shown in Table~\ref{t:both-vs-either}. Another way to view this is
that a higher fraction of the {\em both} papers have \mbox{$r' >
50\%$} than the {\em either} papers.
\begin{table}[t]
\begin{center}
\begin{tabular}{|l|l|l|} \hline
Paper Type & average $r'$ & Fraction with $r' > 50\%$ \\ \hline
{\em both}     & 74\%  & 85\% (41 of 48 papers) \\
{\em either} (but not both) & 35\%  & 19\% (8 of 42 papers)\\ \hline
\end{tabular}
\caption{{\em both} versus {\em either} papers}\label{t:both-vs-either}
\end{center}
\end{table}

The {\em either} papers may be harder because they seem more likely to also
have experimental results that only apply to laboratory-produced
mutants (results not of interest), which can obscure the results that
are of interest (wild-type).

\section{Discussion: Lessons Learned}
\begin{figure*}[t]
\begin{center}
\epsfig{file=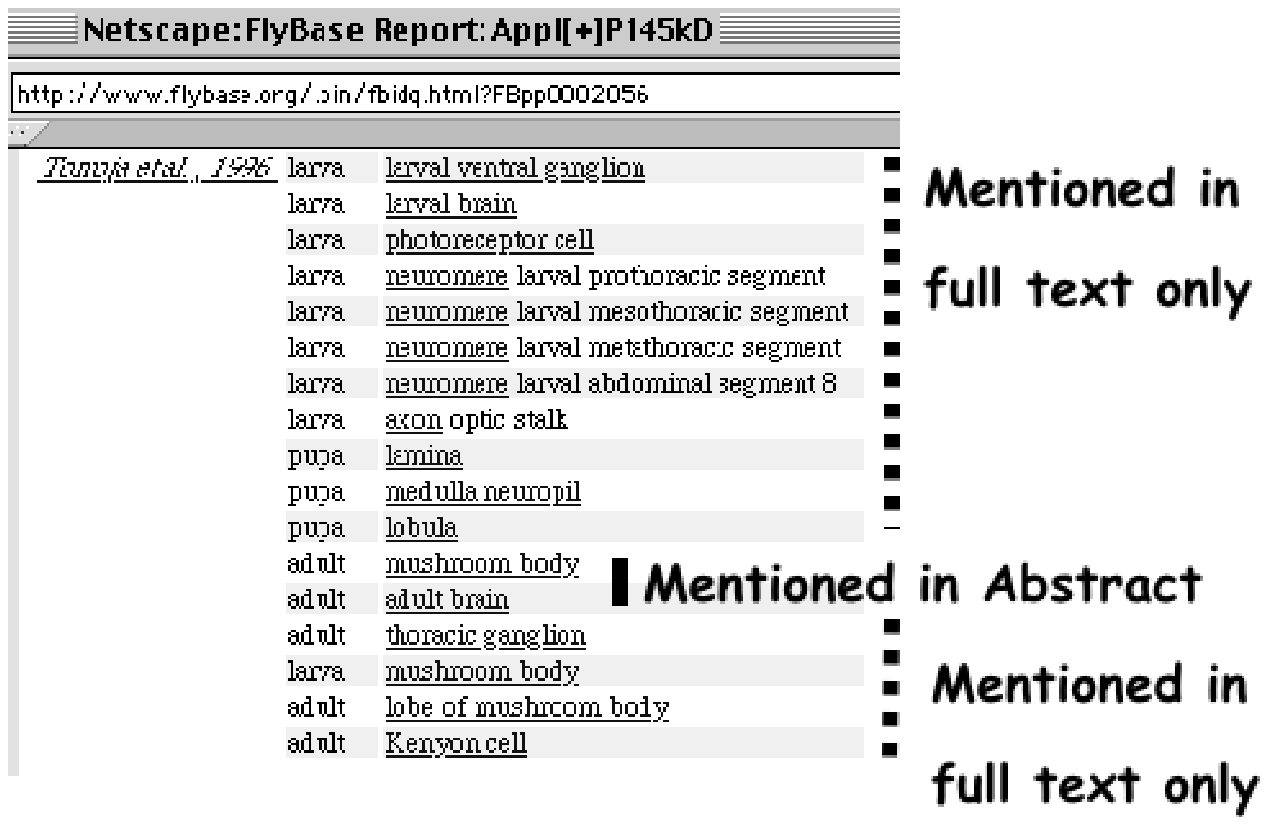}
\caption{Expression patterns found in full text versus abstract}
\label{f:full-vs-abstract}
\end{center}
\end{figure*}

One lesson we learned from running this contest is that access to the
literature itself is a complex matter.  Abstracts of
papers are fairly easy to obtain via PubMed/Medline \\
(http://www.ncbi.nlm.nih.gov/entrez/query.fcgi). However, many of the
results of interest to the FlyBase curators are only described in the
full paper, and not in the abstract. As an example, for the protein
\mbox{\bf Appl$^{+}$P145kD}, FlyBase records that
\citep{Torroja96} finds 17 expression patterns relating to {\it when}
(in the life cycle) and {\it where} (in the body) that protein is
found. Only 2 (12\%) of these patterns (an adult's brain and an
adult's mushroom body) are mentioned in that paper's abstract. The
other 15 (88\%) patterns (for example: a larva's photoreceptor cell
and a pupa's lobula) are only mentioned in the full paper -- see Figure~\ref{f:full-vs-abstract}.

Using full papers introduces complications.  One complication is that
easily accessible electronic versions of some papers do not exist.
Other papers could be obtained in PDF format, but they were not
suitable for processing by most text mining systems.
A subset of the papers were available in HTML format; however, this
HTML version needed to be freely available to the public.  For the
contest, we began with a list of 7100 possible papers, but were able
to obtain only 1118 freely available papers in HTML, from which both the training and
test papers were drawn.

HTML has its own challenges. Publishers set up many of the
HTML versions of the papers so that the main file and the directory to the
linked files have the same path name. The linked files include most
figures and also some figure captions and tables. So in a
straightforward download, one cannot get both the main file and the
linked files. Either one downloads the main file first and then
replaces it with the directory when a linked file is downloaded, or
vice-versa. We chose to keep the main files and leave out the
directory and associated linked
files.  FlyBase curators have mentioned that many of the experimental
results are presented in figures and their captions (B.Matthews,
personal communication). Fortunately, most captions were not left
out, and the captions typically described what was of interest in the
actual figure. Also, most text processing systems cannot actually
handle the figures (images) themselves.

Another complication was that many automated text processing systems
have been designed to handle plain text, but, as mentioned in {\bf The
training and test data} Section, full papers of interest to FlyBase
often have information expressed in typesetting conventions, such as
superscripts, subscripts, italics and Greek letters. It was necessary
to apply conversion routines to produce versions that translated
typographic conventions into plain text corresponding to the FlyBase
conventions.

A second lesson is that the information of interest can differ quite a
bit in appearance between the paper and the corresponding curated
database entries - for example, in gene product naming (see {\bf The
training and test data} Section). FlyBase does not store pointers
to the specific passage(s) that support a database entry. As a result,
finding the evidence for a given entry may require significant biology
expertise and sometimes, also expertise in FlyBase conventions.

This competition was held at a data-mining conference, so not
surprisingly, many contestants made use of statistical, automated
learning and/or automated weighting techniques, including Na\"{\i}ve
Bayes, SVMs, Widrow Hoff linear classifiers, linear regression and the
`tf-idf' weighting scheme. However, such techniques were not enough
to do well. The contestants also needed to either manually determine
what features to look for and/or where to look for them. Examples of
features included keywords, patterns of words and
types of patterns. 
The winning team used manually determined patterns, while
the honorable mention teams used mixtures of manually determined items
and items gleaned from statistics or automated learning.

It was also important to know {\em where} to look for features or patterns.
The winning system made good use of information in figure captions. 
A number of 
groups looked only at sentences containing gene name(s) or certain
keyword(s). Some groups made use of the document structure,
preferring to look in certain sections (e.g., `Results' or
`Methods') and avoiding
other sections. The `References' section is one to especially avoid,
as it contains citations that included
names of genes not discussed in the paper.

The third sub-task was the hardest and the most fine-grained. This
sub-task required determining which genes in a paper had experimental
data on their wild-type (non-mutant) products, as opposed to just
making an overall determination for the paper. So especially for this
sub-task, a contestant's system needed to do more than look for good
indicators of experimental results and good indicators of results
for wild-type versus mutant genes. 
The system also needed to associate indicator terms relating to experimental
findings (e.g., `Northern blot' or `Western blot')
with particular genes.
Some of the high performing systems handled this by looking for
particular patterns of words that would associate an indicator with a
particular gene, with the patterns often being contained within a
sentence or two. Another system handled this by measuring how close
(in sentences) an indicator (feature) was to a gene name and
restricting the measurements to occurrences of gene and indicator
mentions within the same paragraph. 

A common feature of these approaches was that they used
information about the 
document structure and linguistic
structure of a paper, e.g., sections, paragraphs,
sentences, and phrases. This
is in contrast to the information retrieval approach of treating a
paper as just an unstructured set of words.  
We expect that systems will need to make more extensive use of
linguistic and document structure to achieve better results
and to accommodate more realistic tasks.  For example,
linguistic structure may provide critical clues once
the simplifications mentioned in
{\bf The training and test data} Section are removed,
including requiring systems to handle mentions of foreign genes, 
lethal genes or anonymous genes. Similarly, if the list of genes
is not provided in advance for each paper, 
this makes the task
of identifying the set of genes discussed in an article
more difficult.
The system would have to
determine when a new name refers to a new gene and when it is a
synonym for something already known. In this case too,
both linguistic structures
and document structures can provide critical information 

One of our goals in running this evaluation was to evaluate the
evaluation.  For this, we defined three criteria:
\begin{itemize}
\item The evaluation should be repeatable and affordable. This 
should include a reusable training data set, cost-effective preparation
of `gold standard' data for test and repeatable scoring
procedures that are easy to run and easy to understand.
\item The evaluation must attract participants. This means that
it needs to be a problem of importance to biologists, but
also accessible to researchers in text mining teamed with biologists.
\item The task must be tractable but should also push the state of the art.
If the task is well chosen, groups will demonstrate that they are
on the path to the development of a useful capability.
\end{itemize}

Our assessment of the KDD evaluation was that it was successful
along all of these dimensions.  It was affordable.
We estimate that it took us approximately 9 staff months of time to
complete the tasks associated with setting up and running the KDD
evaluation, including: (1) defining the task; (2) obtaining and
normalizing the texts; (3) preparing and packaging the training data;
(4) releasing the training data and answering questions; (5)
developing and explaining the scoring routines; and (6) scoring
the test results.
In addition to our time, it took 2 staff months of time from the
FlyBase curators to curate the test set and answer questions (both our
questions and those of the participants).

We were able to create a reusable training corpus, which we will
continue to make available.\footnote{\hspace{1ex}To obtain the training corpus,
send e-mail to Alex Yeh, asy@mitre.org}

We were able to attract a reasonable number of participating
groups (18) from a wide range of countries.  However, because
of the venue (KDD), we attracted mostly researchers in text mining,
rather than biologists. We would like to attract more participation
from the biology community.

The task we chose is one of real importance to curators responsible
for maintaining biological databases. We believe that there are
many other text mining tasks that could be of great potential
utility to biological database curators.

\section{Conclusions}

We successfully organized 
an initial evaluation on text mining systems to aid
biological database curation, as part of the KDD Challenge Cup 2002. 
Many teams took part in the evaluation, and
their results indicate that curated data
from a biological database can be used to train text mining systems 
to perform a potentially useful task.

The task that we presented to the contestants is only a small part of what
the FlyBase Harvard curators do. But even this limited task is of real
importance to the curators, because most of the papers (for example,
2/3 of our training papers) given to the curators contain no results
of interest, and filtering out such papers is useful. The results from
the ranked-list sub-task look especially promising (the best teams
were 81\%-84\%). But we need to perform further experiments to 
determine whether the
resulting lists will actually help the curators with filtering papers.

We are now involved in planning a larger competition, together with
A. Valencia and C. Blaschke (CNB-Madrid), 
under the umbrella of the ISCB BioLINK Special Interest Group for
Text Data Mining 
(see http://www.pdg.cnb.uam.es/BioLINK/).
We are planning two tasks; the first is the extraction
of gene or protein names from text, so that we can evaluate the
current state of the art in biological entity extraction across systems
that have been reported in the literature over the past few years.
The second task
will require systems 
to associate Gene Ontology
(GO) terms with mentions of proteins in articles curated in the
SWISS-PROT database.  Our experience in organizing the KDD competition
leads us to believe that by using data from curated
databases and focusing on tasks of
immediate utility both to database curators and to researchers,
we can define a good challenge evaluation
for text data mining systems.

\section{Acknowledgements}
This paper
reports on work done in part at the MITRE Corporation under the
support of the MITRE Sponsored Research Program. In addition, many
people at FlyBase have contributed to the KDD Cup task,
especially William Gelbart, Beverly Matthews, Leyla Bayraktaroglu,
David Emmert and Don Gilbert.


\begin{thebibliography}{99}

\bibitem[Duda \etal(2001)]{Duda01}
\begin{book}
\snm{Duda}
\fnm{R.O.}
\init{}
\snm{Hart}
\fnm{P.E.}
\init{}
\snm{Stork}
\fnm{D.G.}
\init{}
\yr{2001}
\bktitle{Pattern Classification}
\orgname{John Wiley \& Sons}
\city{New York}
\end{book}
\hspace{1em}2nd edition.

\bibitem[FlyBase Consortium(2002)]{FlyBase02}
\begin{article}
\snm{FlyBase Consortium}
\fnm{}
\init{}
\yr{2002}
\atitle{The flybase database of the {\it Drosophila} genome projects and community literature}
\jtitle{Nucleic Acids Res.}
\vol{30}
\fpage{106}
\lpage{108}
\end{article}
\hspace{1em}http://flybase.org/

\bibitem[Ghanem \etal(2003)]{Ghanem03}
\begin{article}
\snm{Ghanem}
\fnm{M.M.}
\init{}
\snm{Guo}
\fnm{Y.}
\init{}
\snm{Lodhi}
\fnm{H.}
\init{}
\snm{Zhang}
\fnm{Y.}
\init{}
\yr{2003}
\atitle{Automatic scientific text classification using local patterns: {KDD C}up 2002 (task 1)}
\jtitle{SIGKDD Exploration {\rm newsletter}}
\vol{4 (2)}
\fpage{95}
\lpage{96}
\end{article}
\hspace{1em}http://www.acm.org/sigkdd/explorations/

\bibitem[Hirschman(1998)]{Hirschman98}
\begin{article}
\snm{Hirschman}
\fnm{L.}
\init{}
\yr{1998}
\atitle{The evolution of evaluation: Lessons from the message understanding conferences}
\jtitle{Computer Speech and Language}
\vol{12}
\fpage{281}
\lpage{305}
\end{article}

\bibitem[Hirschman \etal(2002)]{Hirschman02}
\begin{article}
\snm{Hirschman}
\fnm{L.}
\init{}
\snm{Park}
\fnm{J.C.}
\init{}
\snm{Tsujii}
\fnm{J.}
\init{}
\snm{Wong}
\fnm{L.}
\init{}
\snm{Wu}
\fnm{C.H.}
\init{}
\yr{2002}
\atitle{Accomplishments and challenges in literature data mining for biology}
\jtitle{Bioinformatics} 
\vol{18}
\fpage{1553}
\lpage{1561}
\end{article}

\bibitem[Kolhekar \etal(1997)]{Kolhekar97}
\begin{article}
\snm{Kolhekar}
\fnm{A.S.}
\init{}
\snm{Roberts}
\fnm{M.S.}
\init{}
\snm{Jiang}
\fnm{N.}
\init{}
\snm{Johnson}
\fnm{R.C.}
\init{}
\snm{Mains}
\fnm{R.E.}
\init{}
\snm{Eipper}
\fnm{B.A.}
\init{}
\snm{Taghert}
\fnm{P.H.}
\init{}
\yr{1997}
\atitle{Neuropeptide amidation in {\it Drosophila}: Separate genes encode the two enzymes catalyzing amidation}
\jtitle{J. Neurosci.} 
\vol{17}
\fpage{1363}
\lpage{1376}
\end{article}

\bibitem[Regev \etal(2003)]{Regev03}
\begin{article}
\snm{Regev}
\fnm{Y.}
\init{}
\snm{Finkelstein-Landau}
\fnm{M.}
\init{}
\snm{Feldman}
\fnm{R.}
\init{}
\yr{2003}
\atitle{Rule-based extraction of experimental evidence in the biomedical domain - the {KDD C}up 2002 (task 1)}
\jtitle{SIGKDD Explorations {\rm newsletter}}
\vol{4 (2)}
\fpage{90}
\lpage{92}
\end{article}
\hspace{1em}http://www.acm.org/sigkdd/explorations/

\bibitem[Rosen \etal(1989)]{Rosen89}
\begin{article}
\snm{Rosen}
\fnm{D.R.}
\init{}
\snm{Martin-Morris}
\fnm{L.}
\init{}
\snm{Luo}
\fnm{L.}
\init{}
\snm{White}
\fnm{K.}
\init{}
\yr{1989}
\atitle{A {\it Drosophila} gene encoding a protein resembling the human $\beta$-amyloid protein precursor}
\jtitle{Proc. Natl. Acad. Sci. USA}
\vol{86}
\fpage{2478}
\lpage{2482}
\end{article}

\bibitem[Salton \& McGill(1983)]{Salton83}
\begin{book}
\snm{Salton}
\fnm{G.}
\init{}
\snm{McGill}
\fnm{M.J.}
\init{}
\yr{1983}
\bktitle{Introduction to Modern Information Retrieval}
\orgname{McGraw-Hill Book Co.}
\city{New York}
\end{book}

\bibitem[Shi \etal(2003)]{Shi03}
\begin{article}
\snm{Shi}
\fnm{M.}
\init{}
\snm{Edwin}
\fnm{D.S.}
\init{}
\snm{Menon}
\fnm{R.}
\init{}
\snm{Shen}
\fnm{L.}
\init{}
\snm{Lim}
\fnm{J.Y.K.}
\init{}
\snm{Loh}
\fnm{H.T.}
\init{}
\yr{2003}
\atitle{A machine learning approach for the curation of biomedical literature--{KDD C}up 2002 (task 1)}
\jtitle{SIGKDD Explorations {\rm newsletter}}
\vol{4 (2)}
\fpage{93}
\lpage{94}
\end{article}
\hspace{1em}http://www.acm.org/sigkdd/explorations/

\bibitem[Tingvall \etal(2001)]{Tingvall01}
\begin{article}
\snm{Tingvall}
\fnm{T.\"{O}.}
\init{}
\snm{Roos}
\fnm{E.}
\init{}
\snm{Engstr\"{o}m}
\fnm{Y.}
\init{}
\yr{2001}
\atitle{The {GATA} factor Serpent is required for the onset of the humoral immune response in {\it Drosophila} embryos}
\jtitle{Proc. Natl. Acad. Sci. USA}
\vol{98}
\fpage{3884}
\lpage{3888}
\end{article}

\bibitem[Torroja \etal(1996)]{Torroja96}
\begin{article}
\snm{Torroja}
\fnm{L.}
\init{}
\snm{Luo}
\fnm{L.}
\init{}
\snm{White}
\fnm{K.}
\init{}
\yr{1996}
\atitle{{APPL}, the {\it Drosophila} member of the {APP}-family, exhibits differential trafficking and processing in {CNS} neurons}
\jtitle{J. Neurosci.}
\vol{16}
\fpage{4638}
\lpage{4650}
\end{article}

\bibitem[Voorhees \& Buckland(2002)]{Voorhees02}
\begin{book}
\snm{Voorhees}
\fnm{E.M.}
\init{}
\snm{Buckland}
\fnm{L.P., ed.}
\init{}
\yr{2002}
\bktitle{The Eleventh Text REtrieval Conference (TREC 2002)}
\orgname{NIST Special Publication 500-XXX}
\city{Gaithersburg, Maryland, USA}
\end{book}
\hspace{1em}http://trec.nist.gov/pubs/trec11/t11\_proceedings.html

\bibitem[Yeh \etal(2003)]{Yeh03}
\begin{article}
\snm{Yeh}
\fnm{A.}
\init{}
\snm{Hirschman}
\fnm{L.}
\init{}
\snm{Morgan}
\fnm{A.}
\init{}
\yr{2003}
\atitle{Background and overview for {KDD C}up 2002 task 1: Information extraction from biomedical articles}
\jtitle{SIGKDD Explorations {\rm newsletter}}
\vol{4 (2)}
\fpage{87}
\lpage{89}
\end{article}
\hspace{1em}http://www.acm.org/sigkdd/explorations/

\bibitem[Yun \etal(2000)]{Yun00}
\begin{article}
\snm{Yun}
\fnm{B.}
\init{}
\snm{Lee}
\fnm{K.}
\init{}
\snm{Farkas}
\fnm{R.}
\init{}
\snm{Hitte}
\fnm{C.}
\init{}
\snm{Rabinow}
\fnm{L.}
\init{}
\yr{2000}
\atitle{The {LAMMER} protein kinase encoded by the {{\it Doa}} locus of Drosophila is required in both somatic and germline cells and is expressed as both nuclear and cytoplasmic isoforms throughout development}
\jtitle{Genetics} 
\vol{156}
\fpage{749}
\lpage{761}
\end{article}

\end{thebibliography}
\end{document}